# SYSTEMATIC IMPROVEMENT OF USER ENGAGEMENT WITH ACADEMIC TITLES USING COMPUTATIONAL LINGUISTICS

| | | |
|---|---|---|
| Nim Dvir * | State University of New York at Albany, Albany, NY, USA | Ndvir@albany.edu |
| Ruti Gafni | The Academic College of Tel Aviv Yaffo, Tel Aviv, Israel | rutigafn@mta.ac.il |

* Corresponding author

## ABSTRACT

| | |
|---|---|
| Aim/Purpose | This paper describes a novel approach to systematically improve information interactions based solely on its wording. |
| Background | Providing users with information in a form and format that maximizes its effectiveness is a research question of critical importance. Given the growing competition for users' attention and interest, it is agreed that digital content must engage. However, there are no clear methods or frameworks for evaluation, optimization and creation of such engaging content. |
| Methodology | Following an interdisciplinary literature review, we recognized three key attributes of words that drive user engagement: (1) Novelty (2) Familiarity (3) Emotionality. Based on these attributes, we developed a model to systematically improve a given content using computational linguistics, natural language processing (NLP) and text analysis (word frequency, sentiment analysis and lexical substitution). We conducted a pilot study (n=216) in which the model was used to formalize evaluation and optimization of academic titles. A between-group design (A/B testing) was used to compare responses to the original and modified (treatment) titles. Data was collected for selection and evaluation (User Engagement Scale). |
| Contribution | The pilot results suggest that user engagement with digital information is fostered by, and perhaps dependent upon, the wording being used. They also provide empirical support that engaging content can be systematically evaluated and produced. |
| Findings | The preliminary results show that the modified (treatment) titles had significantly higher scores for information use and user engagement (selection and evaluation). |
| Recommendations for Practitioners | We propose that computational linguistics is a useful approach for optimizing information interactions. The empirically based insights can inform the development of digital content strategies, thereby improving the success of information interactions. |
| Recommendations for Researchers | By understanding and operationalizing content strategy and engagement, we can begin to focus efforts on designing interfaces which engage users with features appropriate to the task and context of their interactions. This study will benefit the information science field by enabling researchers and practitioners alike to understand the dynamic relationship between users, computer applications and tasks, how to assess whether engagement is taking place and how to design interfaces that engage users. |
| Impact on Society | This research can be used as an important starting point for understanding the phenomenon of digital information interactions and the factors that promote and facilitates them. It can also aid in the development of a broad framework for systematic evaluation, optimization, and creation of effective digital content. |
| Future Research | Moving forward, the validity, reliability and generalizability of our model should be tested in various contexts. In future research, we propose to include additional linguistic factors and develop more sophisticated interaction measures. |
| Keywords | information behavior, text analysis, computational linguistics, information interaction, user experience (UX), knowledge acquisition, decision-making, user engagement, content strategy, digital nudging |

## INTRODUCTION

Providing users with information in a form and format that maximizes its effectiveness is a research question of critical importance. With increasing amount of digital content being published by commercial businesses, governments, healthcare organizations and private citizens, information interactions and information delivery become more complex (Toms, 2002; Wilson, 2000).

In recent years there has been a shift from usability and effectiveness of the information to a focus on user experience (UX). The international standard on ergonomics of human system interaction defines UX as "a person's perceptions and responses that result from the use or anticipated use of a product, system or service". User engagement (UE) is a quality of UX broadly defined as users' level of cognitive, psychological and behavioral involvement and/or investment when interacting with a technological resource (O'Brien & Cairns, 2016; O'Brien & Toms, 2008). UE has been increasingly used to describe successful user interactions with technology. Engagement is considered a desirable - even essential - human response to computer-mediated activities (Dvir, 2019; O'Brien, 2018; Sutcliffe, 2016). Given the competition for users' attention and interest, it is agreed that technologies must engage users to be successful.  It is believed that the ability to engage and sustain engagement in digital environments can result in positive outcomes for citizen inquiry and participation, e-health, web search, e-learning, and so on (Kostkova, 2016; O'Brien & Cairns, 2016). Yet, UE is an abstract construct that manifests differently within different computer-mediated contexts. This has made it challenging to define, design for, and evaluate. Current research is unstructured and spread across various disciplines, leading to wide-ranging, and sometimes disparate, perspectives, vocabularies, and measurement methodologies.

A main research question is how to design information systems to support UE. Designing for engaging experiences is an oft-cited goal of interactive system development in many disciplines, yet there are no guidelines to channel designers' efforts to make things engaging (Blythe, Monk, Overbeeke, & Wright, 2005; Overbeeke, Djajadiningrat, Hummels, Wensveen, & Prens, 2003).

This paper argues that the way information is delivered by the system user interface (UI) plays a large part in how and when it is used, engaged with and experienced. This is called *digital nudging* - the employment of UI design elements to guide people's behavior in digital choice environments (Weinmann, Schneider, & Brocke, 2016). Specifically, based on principles from the Cumulative Prospect Theory (CPT) and framing effect (Tversky & Kahneman, 1992), we suggest that the phrasing of digital information - the words being used - can impact how the information will be consumed, perceived and used.

Following a comprehensive literature review, we propose a conceptual and practical framework to evaluate and improve a given content based on the identification of reliable and reusable metrics and employment of textual analysis and computational linguistics. We conceptualize and operationalize UE creation by identifying three key attributes of digital content that drive successful information interactions: (1) Familiarity - words that are known and popular, operationalized as having high frequency in popular culture (cultural relevance); (2) Novelty - words that are rare in the context of the interaction (operationalized as having low frequency in that context); (3) Emotionality - emotive words that evoke emotional reaction, operationalized through sentiment polarity (positive, negative, or neutral).

We call these words "sticky words." We hypothesized that when strategically placed in a given content, these words can increase the motivation to consume it, improve how it is evaluated and lead to better knowledge retention. We conducted a pilot study in which our model was used to formalize evaluation and optimization of academic titles.

The following section describe our theoretical underpinning, research model and pilot study.

## LITERATURE REVIEW

The Merriam-Webster dictionary defines the term engagement as a "state of emotional involvement or commitment" and "process of drawing favorable attention and interest" ("Engagement," 2018). Both as a state and a process, engagement has been considered a desirable goal in various contexts, such as government, education, marketing, health and more. For example, in the context of government and public policy, the terms the "civic engagement" or "political engagement" indicate healthy participation in the political process (Dvir, 2017; Navajas, 2014; Obar, Zube, & Lampe, 2011). In business and marketing, "consumer engagement" is used frequently to describe positive consumer relationships with a company or a brand (Chu & Kim, 2011; Dvir & Gafni, 2018).

User engagement is a term used to describe the expression of the interactions with information and communication technologies (ICT). Most research on UE emphasizes user-centered approaches, highlighting the individual's affective, behavioral and cognitive factors of human-information interaction, and the need for technologies that stimulate each of these components (O'Brien & Cairns, 2016).

O'Brien and Toms (2008) developed the Process model of UE, which focuses on the relationships amongst the variables related to information interaction with the goal of predicting or identifying outcomes. The model suggests that UE is not a singular element but rather a process, as users move in the same trajectory when interacting with ICT. This conceptual behavioral model views UE as a process in which computer users initiate and sustain engagement, disengage with the application or task, and potentially re-engage once or several times. The model is comprised of four distinct stages: point of engagement, period of sustained engagement, disengagement, and re-engagement. Each of these is characterized by a set of attributes.

**The point of engagement** occurs when the user decides to invest in the interaction. It is initiated by the aesthetic appeal and novelty of the interface, interest, motivation, or a specific or experiential goal to be achieved through the interaction.

**Period of sustained engagement** happens when users are able to maintain their attention and interest in the application.

**Disengagement** occurred when participants made an internal decision to stop the activity, or when factors in the participants' external environment caused them to cease being engaged.

**Re-engagement** is returning to an application in future as the result of positive experience with that application.

By segmenting UE into phases, O'Brien & Toms were able to identify attributes that seemed most salient for that phase of the interaction and therefore predict and facilitate engagement. As a result, O'Brien and Toms developed an attribute-based questionnaire called the User Engagement Scale (UES), which is used as a methodological framework for measurement UE with a technology (O'Brien & Cairns, 2015; O'Brien, Cairns, & Hall, 2018; O'Brien & Toms, 2013). The UES consists of a set of questions, which tap into various dimensions of UE: novelty and aesthetics reflect users' attraction to the media system or interface, focused attention and felt involvement capture cognitive and emotional focus on media content and the endurability of system use represents evaluations of success and voluntary participation to recommend the website to others. In further work, the UES was refined to a new short form, in which the items did not change but were grouped differently. The modified scale includes questions relating to aesthetic appeal, focused attention, perceived usability and reward and endurability (O'Brien et al., 2018). Table 1 lists the various attributes used in the UES:

**Table 1. UES Attributes and definitions**

| Attributes | Definition |
| --- | --- |
| **Novelty and Aesthetic Appeal** | Users' level of interest and curiosity evoked by the system and its contents. The users' perception of the visual appearance of a computer application interface |
| **Focused Attention** | The concentration of mental activity; contains some elements of Flow, specifically focused concentration, absorption, and temporal dissociation |
| **Felt Involvement** | Users' feelings of being drawn in, interested, and having fun during the interaction |
| **Perceived Usability** | Users' affective (e.g., frustration) and cognitive (e.g., effort) responses to the system |
| **Endurability and reward** | Users' overall evaluation of the experience, its perceived success and whether users would recommend the site to others. This factor combines concepts related to users' likelihood to return and evaluation of system success. |

The UE Process Model and the UES have been tested for reliability, validity, and generalizability in various domains (online news, shopping, digital games, social media). These studies corroborate the UE process model and the feasibility of a universal measure of UE (O'Brien & Cairns, 2016;

O'Brien & McKay, 2018). The studies also demonstrated that UE is consistent across diverse types of applications. The UES has been adopted by more than 50 international research teams, who have used it to examine UE with educational technologies, search systems, haptic technologies, health and consumer applications (O'Brien & Cairns, 2016; O'Brien & McKay, 2018).

However, the model has a few gaps: first, engagement with information at a theoretical and conceptual level is still limited. For example, usability attributes such as effectiveness, efficiency, and satisfaction are intricately woven into the experience of engagement; while an application may be usable, it may not be engaging, but engaging applications do appear to have an inherent baseline of usability (O'Brien & Toms, 2008). This also includes ease-of-use and flow, as the perceived effort is suggested to be a barrier to engagement (O'Brien & McKay, 2018).

Second, system design is not fully incorporated. System features have been empirically shown to drive desirable user behavior. Designing for engaging experiences is an oft-cited goal of interactive system development in many disciplines, yet there are no guidelines to channel designers' efforts to make things engaging (Blythe et al., 2005; Overbeeke et al., 2003). **In particular,** the way information is delivered through the UI was demonstrated to be a crucial part of UE. This also includes digital content, defined as the textual or visual information made available by a website or other electronic medium (Rowley, 2008). In marketing and online commerce, digital content and digital information are often synonymous terms.

O'Brien (2011), O'Brien, Freund, and Westman (2014), and O'Brien, Freund, and Kopak (2016) found that content was an important quality of users' engagement when interacting with ICT. Thus, they suggested that the user engagement framework be broadened to incorporate content, and for more investigation of the relationship between the nature of the content, how it is presented through the interface and its effectiveness and success (O'Brien et al., 2014; O'Brien, 2011, 2016; O'Brien & McKay, 2016).

Several aspects of content have been investigated in conjunction with UE, including sentiment, interest, novelty, quality and message framing (O'Brien, 2017). Digital content has been proven to play a significant role in influencing consumers' behavior online (Gafni & Dvir, 2018). For example, the amount of content provided on landing pages – standalone web pages created explicitly for marketing or advertising campaigns - was proven to impact consumers' behavior and willingness to disclose personal data (Dvir & Gafni, 2018). Despite current research, a shared understanding of how to usefully conceptualize engaging content is lacking. Current research is focused on identifying and evaluating UE, but not on the actual development of engaging experiences (O'Brien, 2017; O'Brien & Cairns, 2016). Content is not well situated in current theory and not thoroughly researched as a determinant of UE. The UE model and scale do not emphasize content; Rather, they focused primarily on the characteristics of the system and how these were perceived and acted upon by users (O'Brien, 2011, 2017).

# Cumulative Prospect Theory

One of the biggest challenges in understanding human information interactions is that much of them occur at a subconscious level. Research in the area of embodied cognition offers a powerful and effective way to better understand how user interact with information. Particularly, the drivers behind human judgment and decision making.

Cumulative prospect theory (CPT) is a model for descriptive decisions under risk and uncertainty which was developed by Amos Tversky and Daniel Kahneman (Kahneman & Tversky, 1979; Tversky & Kahneman, 1981, 1992). A key principle of CPT (and its predecessor Prospect Theory) is that people tend to think of possible outcomes relative to a certain reference point (often the status quo), rather than to the final status. This phenomenon is called "framing effect". It suggests that the context (or framing) has great impact on decision-making. Therefore, information interactions can be "nudged" using extrinsic manipulation of the decision-options offered, as well as from forces intrinsic to decision-makers. Prospect theory distinguishes two phases in the choice process: framing and valuation. In the framing phase, the decision maker constructs a representation of the acts, contingencies, and outcomes that are relevant to the decision. In the valuation phase, the decision maker assesses the value of each prospect and chooses accordingly .The theory establishes that selection is impacted by "heuristics", which are simple, efficient rules used to form judgments and make decisions. Heuristics are "mental shortcuts" which often lead to focusing on one aspect of a complex problem and ignoring others. Heuristics can work well under most circumstances, but they can also cause systematic deviations from logic, probability or rational choice. The resulting errors are called "cognitive biases" - unconscious, automatic influences on judgment and decision making that reliably produce reasoning errors.

# RESEARCH MODEL

While the UE theory and models provide an important basis, there is a need for the introduction of more sophisticated predictors (such as content attributes) and digital nudging techniques (such as textual analysis and computational linguistics). The UE process model does not emphasize content; Rather, it focused primarily on the characteristics of the system and how these were perceived and acted upon by users.

The objective of this research is to develop a framework for systematic evaluation, optimization, and creation of engaging content interactions. This paper is guided by the following research questions:

## R1: CAN THE PHRASING OF DIGITAL INFORMATION – THE WORDS BEING USED – IMPACT UE?

Based on the Cumulative Prospect Theory (CPT) and the framing effect, we suggest that the context (or framing) of the information delivered to users can assist in extrinsic manipulation of their interactions. Specifically, the words being used to deliver the message can impact UE.

We identify three extrinsic attributes of digital content of UE, which we call "Sticky words" (Figure 1):

- **Novelty –** novelty is described in the UES as users' level of interest and curiosity evoked by the system and its contents. We operationalized it as words that are rare in the context of the interaction (distinctive), operationalized as having low frequency in that context.
- **Familiarity –** The UES also emphasizes perceived usability and accessibility. Therefore, we suggest that the words should also be known and popular, operationalized as having high frequency in popular culture (cultural relevance).
- **Emotionality –** The UES suggests that engagement is evoked by an emotional reaction, which stimulates voluntary, focused attention, interest and enjoyment. We operationalized it as emotive words, recognized through sentiment polarity (positive or negative).

We call these words "sticky words." We hypothesized that when strategically placed in a given content, these words can impact UE with content. Specifically, increase the motivation to use it (selection) and improve how it is experienced (evaluation).

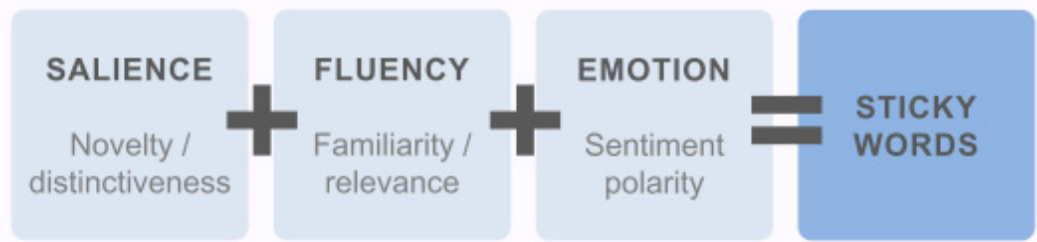

**Figure 1. The "sticky words" model**

We also argue that these words can be systematically recognized using word frequency, sentiment analysis and semantic substitution.

## R2. CAN UE WITH INFORMATION BE SYSTEMATICALLY EVALUATED, CREATED OR MANIPULATED?

Based on the previous question, we suggest the UE can be manipulated based on the wording used (its delivery). We argue that these words can be employed to evaluate and improve any given content based on reliable and reusable metrics, and that it can be done systematically using textual analysis and computational linguistics. Specifically, we suggest that UE can be developed using the following techniques:

• **Lexical substitution –** refers to the task of identifying meaning-preserving substitutes for a targeted term in a given sentential context, and replacing them without altering the meaning of the sentence.

• **Term frequency–inverse document frequency (TF–IDF) -** a central tool in scoring and ranking content relevance. Effective TF-IDF is reached by the combination of high term frequency (in the given setting) and a low document frequency of the term in the whole collection of documents.

• **Sentiment analysis –** an automatic measure of textual emotional polarity.

## METHOD

We conducted a pilot study in which our model was used to formalize evaluation and optimization of academic titles. A randomized between-group design was employed to observe how changes in wording (independent variable) impact UE. We measured it in two ways: (a) Behavioral (information use), operationalized as selection to read the article (b) Experiential (affect and perception), operationalized as self-reported evaluation of the title (using the questions included in the UES).

We developed a computational linguistic process for compiling a dataset (Figure 2). We collected academic titles from the JSTOR database. We created a corpus of potential "sticky words" using the movie keyword analyzer of The Internet Movie Database (IMDB.com), which aggregates all keywords assigned to movies. These represent well-known words frequently used in popular culture. We used Term Frequency–Inverse Document Frequency (TF-IDF) to find words with high frequency on the list of IMDB keywords (for familiarity), and with low frequency in the collection of the academic titles (for novelty). Sentiment analysis was used to categorize the words for emotional polarity (positive, negative, or neutral). We then used lexical and semantic substitution to find synonyms and make replacements of words in the academic titles, replacing only one word at a time with an equivalent

"sticky word." For example, the title "End of the library: Organized information and digital ubiquity" became "Death of library: Organized information and digital ubiquity".

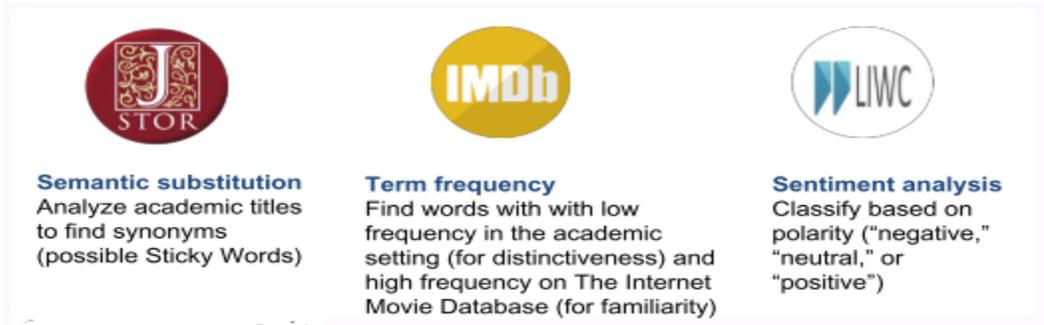

**Figure 2. The computational linguistic process of compiling a dataset**

We manually verify that the introduction of the new word does not alter the meaning of the text. The result is a dataset of original and treatment (modified) titles. Examples are shown in Table 2.

**Table 2. Example of sentence variants tested in this study**

| ORIGINAL | TREATMENT |
|---|---|
| **End** of the library: does digital ubiquity endangers traditional channels of organized information? | **Death** of the library: does digital ubiquity endangers traditional channels of organized information? |
| **Reproductive** activity and the lifespan of male fruit flies | **Sexual** activity and the lifespan of male fruit flies |
| Be a civic **leader**: how to effectively use open data for digital government | Be a civic **hero**: how to effectively use open data for digital government |

Participants were recruited using email sent to a listserv of undergraduate students in a large research university in the U.S. We used the software Qualtrics to randomly present the participants with the titles (either original or treatment) and to collect responses. The interaction was assessed based on dimensions recognized by the UE process model:

(a) Behavior / voluntary participation (selection) - Participants were asked whether they would like to read the title and if so they had to click on it.

(b) Experience / perception (content evaluation) - Participants were asked to rate the title using questions from the User Engagement Scale (UES) (O'Brien et al., 2018).

## RESULTS

Analysis of the results (n-216) is still in primary stages and requires more work. However, Initial data analysis shows the following insights:

### RQ1: IMPACT OF "STICKY WORDS" ON UE (SELECTION AND EVALUATION)

Figure 3 shows the difference in the selection scores recorded for the original and the treatment titles (with the "sticky words"). The number of selected and not selected documents were similar for the original titles (M=.51, SD=.503), but there was a great difference for the treatment titles, which used the "sticky words"(M=.78, SD=.414), as can be seen in Table 3. Table 4 shows the results of the t-test between the groups.

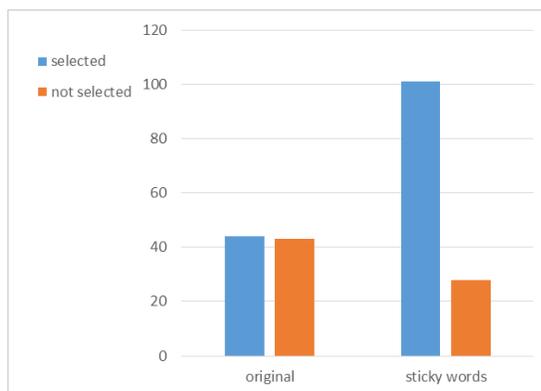

**Figure 3. Selection results for original vs. treatment titles**

**Table 3. Selection statistics**

| Variant | N | Mean | Std. Deviation | Std. Error Mean |
|---|---|---|---|---|
| **original** | 87 | .51 | .503 | .054 |
| **sticky words** | 129 | .78 | .414 | .036 |

**Table 4. Selection samples t-test**

| F | Sig. | t | df | Sig. (2-tailed) | Mean Difference | Std. Error Difference |
|---|---|---|---|---|---|---|
| 40.555 | .000 | 4.423 | 214 | .000 | -.277 | .063 |

There was also a significant difference in the evaluation score (UES average) between the original (M=3.2, SD=1.1) and treatment ("sticky") titles (M=3.8, SD=0.79) conditions as shown in Table 5. The t-test of the evaluation can be seen in Table 6.

**Table 5. Evaluation statistics**

| | Variant | N | Mean | Std. Deviation | Std. Error Mean |
|---|---|---|---|---|---|
| UES | original | 87 | 3.2126 | 1.10530 | .11850 |
| | sticky | 129 | 3.8643 | .79795 | .07026 |

**Table 6. Evaluation Samples Test**

| | | Levene's Test for Equality of Variances | | t-test for Equality of Means | | | | | 95% Confidence Interval of the Difference | |
|---|---|---|---|---|---|---|---|---|---|---|
| | | F | Sig. | t | df | Sig. (2-tailed) | Mean Difference | Std. Error Difference | Lower | Upper |
| UES | Equal variances assumed | 14.545 | .000 | 5.031 | 214 | .000 | -.65170 | .12953 | -.90702 | -.39637 |
| | Equal variances not assumed | | | 4.731 | 145.042 | .000 | -.65170 | .13776 | -.92398 | -.37942 |

Thus, the findings indicate that "sticky words" impact information behavior and lead to more effective interactions. These findings suggest that successful interactions with digital content are fostered by, and perhaps dependent upon, the wording or language being used.

## RQ2: CAN USER ENGAGEMENT BE DEVELOPED SYSTEMATICALLY?

We were able to use term frequency, sentiment analysis and lexical substitution to evaluate and replace "sticky words." The findings indicate the potential to use computational linguistics to identify factors that predict successful interactions. Yet, there is still a need to refine the model to achieve full automation.

## CONCLUSION

The implications of this research are twofold. First, our findings suggest that successful interactions with digital content are fostered by, and perhaps dependent upon, the wording or language being used. Second, we provide empirical support that engaging content can be evaluated and optimized systematically. We propose that computational linguistics is a useful approach for studying online information interactions and that further study can result in a broader conceptualization of content strategy and its evaluation. These empirically based insights can inform the development of digital content strategies, thereby improving the success of information interactions.

This study demonstrates the complex and dynamic relationships between users' behavior, affect and the expression of the information on computer interfaces. By understanding and operationalizing engagement, we can begin to focus efforts on designing interfaces that engage users with features appropriate to the task and context of their interactions. Thus, the contributions of this work are both theoretical and applied. It will benefit the information science field by enabling researchers and practitioners alike to understand the dynamic relationship between users, computer applications and tasks, how to assess whether engagement is taking place and how to design interfaces that engage users.

Moving forward, the validity, reliability and generalizability of our model should be tested in various contexts. In future research, we propose to include additional linguistic factors and develop more sophisticated interaction measures. This research can be used as an important starting point for understanding the phenomenon of digital information interactions and behavior, the factors that promote and facilitates them, and in the development of a broad framework for systematically evaluation, optimization, and creation effective digital content.

## REFERENCES

Blythe, M. A., Monk, A. F., Overbeeke, K., & Wright, P. C. (2005). *Funology from usability to enjoyment*. Kluwer Academic Publishers. Retrieved from http://citeseerx.ist.psu.edu/viewdoc/download?doi=10.1.1.372.6229&rep=rep1&type=pdf

Chu, S. C., & Kim, Y. (2011). Determinants of consumer engagement in electronic word-of-mouth (eWOM) in social networking sites. *International Journal of Advertising*, *30*(1), 47. https://doi.org/10.2501/ija-30-1-047-075

Dvir, N. (2015). *The influence of gender on consumer behavior and decision making in online and mobile learning environments*. Retrieved from https://www.albany.edu/~nd115232/docs/InfluenceOfGender.pdf

Dvir, N. (2017). Mitigating challenges of open government data. *Preprints*. https://doi.org/10.20944/preprints201712.0182.v1

Dvir, N. (2019). What is user engagement? An interdisciplinary perspective on users' interaction with information technology. *Zenodo*. doi: 10.5281/zenodo.2577620

Dvir, N., & Gafni, R. (2018). When less is more: Empirical study of the relation between consumer behavior and information sharing on commercial landing pages. *Informing Science: The International Journal of an Emerging Transdiscipline*, *21*, 19-39. https://doi.org/10.28945/4015

Engagement. (2018). In *Merriam-Webster*. Retrieved from https://www.merriam-webster.com/dictionary/engagement

Gafni, R., & Dvir, N. (2018). How content volume on landing pages influences consumer behavior: Empirical evidence. *Proceedings of the Informing Science and Information Technology Education Conference, La Verne, California*, 035–053. https://doi.org/10.28945/4016

Kahneman, D., & Tversky, A. (1979). Prospect theory: An analysis of decision under risk. *Econometrica*, *47*(2), 263–292. https://doi.org/10.2307/1914185

Kostkova, P. (2016). User engagement with digital health technologies. In H. O'Brien & P. Cairns (Eds.), *Why engagement matters* (pp. 127–156). https://doi.org/10.1007/978-3-319-27446-1_6

Navajas, H. G. de Z. (Ed.). (2014). New technologies and civic engagement; Engaging audiences via online news sites (Ch. 11). In *New Technologies and Civic Engagement: New Agendas in Communication*. New York, NY: Routledge. https://doi.org/10.4324/9781315750927

Obar, J. A., Zube, P., & Lampe, C. (2011). Advocacy 2.0: An analysis of how advocacy groups in the United States perceive and use social media as tools for facilitating civic engagement and collective action. *SSRN Electronic Journal*. https://doi.org/10.2139/ssrn.1956352

O'Brien, H., Freund, L., & Westman, S. (2014). What motivates the online news browser? News item selection in a social information seeking scenario. *Information Research*, *19*(3), 95–112. Retrieved from http://libproxy.albany.edu/login?url=http://search.ebscohost.com/login.aspx?direct=true&db=lih&AN=100304916&site=eds-live&scope=site

O'Brien, H. L. (2011). Exploring user engagement in online news interactions. *Proceedings of the American Society for Information Science and Technology*, *48*(1), 1–10. https://doi.org/10.1002/meet.2011.14504801088

O'Brien, H. L. (2016). Translating theory into methodological practice. In H. L. O'Brien & P. Cairns (Eds.), *Why engagement matters* (pp. 27–52). https://doi.org/10.1007/978-3-319-27446-1_2

O'Brien, H. L. (2017). Antecedents and learning outcomes of online news engagement. *Journal of the Association for Information Science and Technology*, *68*(12), 2809–2820. https://doi.org/10.1002/asi.23854

O'Brien, H. L. (2018). A holistic approach to measuring user engagement. In M. Filimowicz & V. Tzankova (Eds.), *New directions in third wave human-computer interaction: Volume 2 - Methodologies* (pp. 81–102). https://doi.org/10.1007/978-3-319-73374-6_6

O'Brien, H. L., & Cairns, P. (2015). An empirical evaluation of the User Engagement Scale (UES) in online news environments. *Information Processing & Management*, *51*(4), 413–427. https://doi.org/10.1016/j.ipm.2015.03.003

O'Brien, H. L., & Cairns, P. (2016). *Why engagement matters: Cross-disciplinary perspectives of user engagement in digital media*. https://doi.org/10.1007/978-3-319-27446-1

O'Brien, H. L., Cairns, P., & Hall, M. (2018). A practical approach to measuring user engagement with the refined user engagement scale (UES) and new UES short form. *International Journal of Human-Computer Studies*, *112*, 28–39. https://doi.org/10.1016/j.ijhcs.2018.01.004

O'Brien, H. L., Freund, L., & Kopak, R. (2016). Investigating the role of user engagement in digital reading environments. *Proceedings of the 2016 ACM on Conference on Human Information Interaction and Retrieval*, 71–80. https://doi.org/10.1145/2854946.2854973

O'Brien, H. L., & McKay, J. (2016). What makes online news interesting? Personal and situational interest and the effect on behavioral intentions. *Proceedings of the Association for Information Science and Technology*, *53*(1), 1–6. https://doi.org/10.1002/pra2.2016.14505301150

O'Brien, H. L., & McKay, J. (2018). Modeling antecedents of user engagement. In *The handbook of communication engagement* (pp. 73–88). https://doi.org/10.1002/9781119167600.ch6

O'Brien, H. L., & Toms, E. G. (2008). What is user engagement? A conceptual framework for defining user engagement with technology. *Journal of the American Society for Information Science and Technology*, *59*(6), 938–955. https://doi.org/10.1002/asi.20801

O'Brien, H. L., & Toms, E. G. (2013). Examining the generalizability of the User Engagement Scale (UES) in exploratory search. *Information Processing & Management*, *49*(5), 1092–1107. https://doi.org/10.1016/j.ipm.2012.08.005

Overbeeke, K., Djajadiningrat, T., Hummels, C., Wensveen, S., & Prens, J. (2003). Let's make things engaging. In M. A. Blythe, K. Overbeeke, A. F. Monk, & P. C. Wright (Eds.), *Funology* (Vol. 3, pp. 7–17). https://doi.org/10.1007/1-4020-2967-5_2

Rowley, J. (2008). Understanding digital content marketing. *Journal of Marketing Management*, *24*(5–6), 517–540. https://doi.org/10.1362/026725708x325977

Sutcliffe, A. (2016). Designing for user experience and engagement. In H. O'Brien & P. Cairns (Eds.), *Why engagement matters* (pp. 105–126). https://doi.org/10.1007/978-3-319-27446-1_5

Toms, E. G. (2002). Information interaction: Providing a framework for information architecture. *Journal of the Association for Information Science and Technology*, *53*(10), 855–862. https://doi.org/10.1002/asi.10094

Tversky, A., & Kahneman, D. (1981). The framing of decisions and the psychology of choice. *Science*, *211*(4481), 453–458. https://doi.org/10.1126/science.7455683

Tversky, A., & Kahneman, D. (1992). Advances in prospect theory: Cumulative representation of uncertainty. *Journal of Risk and Uncertainty*, *5*(4), 297–323. https://doi.org/10.1007/bf00122574

Weinmann, M., Schneider, C., & Brocke, J. vom. (2016). Digital nudging. *Business & Information Systems Engineering*, *58*(6), 433–436. https://doi.org/10.1007/s12599-016-0453-1

Wilson, T. D. (2000). Human information behavior. *Informing Science, the International Journal of an Emerging Transdiscipline*, *3*(2), 49–56. https://doi.org/10.28945/576

# BIOGRAPHIES

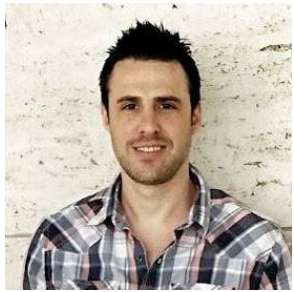

**Nim Dvir** is a Ph.D. candidate in the Information Studies department at the University at Albany. His main research interests are human-computer interaction (HCI), user experience (UX), E-Commerce, Content Strategy, and online user behavior. His work explores the determinants of perception, motivation, engagement, and decision-making in digital environments. He holds a B.A in International Relations and Media Studies from New York University (NYU) and an M.B.A in Marketing and Information Systems from City University of New York (CUNY) Baruch College's Zicklin School of Business, both Cum Laude. For more information: albany.edu/~nd115232/

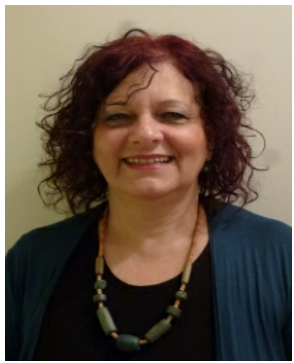

**Ruti Gafni** is the Head of the Information Systems B.Sc. program at The Academic College of Tel Aviv Yaffo. She holds a PhD from Bar-Ilan University, Israel (in the Business Administration School), focusing on Information Systems, an M.Sc. from Tel Aviv University and a BA (Cum Laude) in Economics and Computer Science from Bar-Ilan University. She has more than 40 years of practical experience as Project Manager and Analyst of information systems.